\documentclass[conference]{IEEEtran}
\IEEEoverridecommandlockouts

\usepackage{float}
\usepackage[OT6,T1]{fontenc}
\usepackage[utf8]{inputenc}
\newcommand{\armenian}{\fontencoding{OT6}\fontfamily{cmr}\selectfont}
\DeclareTextFontCommand{\textarmenian}{\armenian}
\usepackage[dvipsnames]{xcolor}
\usepackage{tikz}
\usetikzlibrary{arrows,positioning,shapes,backgrounds,fit}
\usepackage{pgf-pie}
\usepackage{cite}
\usepackage{amsmath,amssymb,amsfonts}
\usepackage{algorithmic}
\usepackage{graphicx}
\usepackage{textcomp}
\usepackage{multirow}
\usepackage{makecell} 
\usepackage[skip=10pt]{caption}

\def\BibTeX{{\rm B\kern-.05em{\sc i\kern-.025em b}\kern-.08em
    T\kern-.1667em\lower.7ex\hbox{E}\kern-.125emX}}

\newcounter{featureSetCounter}

\makeatletter
\def\ps@IEEEtitlepagestyle{%
  \def\@oddfoot{\mycopyrightnotice}%
  \def\@evenfoot{}%
}
\def\mycopyrightnotice{%
  {\footnotesize \begin{minipage}{\textwidth}© 2018 IEEE. Personal use of this material is permitted. Permission from IEEE must be obtained for all other uses, in any current or future media, including reprinting/republishing this material for advertising or promotional purposes, creating new collective works, for resale or redistribution to servers or lists, or reuse of any copyrighted component of this work in other works.\end{minipage}\hfill}
  \gdef\mycopyrightnotice{}
}

\begin{document}
\bstctlcite{IEEEisprasopen:BSTcontrol}

\title{pioNER: Datasets and Baselines for Armenian Named Entity Recognition}

\author{
\IEEEauthorblockN{Tsolak Ghukasyan$^{1}$, Garnik Davtyan$^{2}$, Karen Avetisyan$^{3}$, Ivan Andrianov$^{4}$}
\IEEEauthorblockA{Ivannikov Laboratory for System Programming at Russian-Armenian University$^{1,2,3}$, Yerevan, Armenia\\
Ivannikov Institute for System Programming of the Russian Academy of Sciences$^{4}$, Moscow, Russia\\
Email: \{$^{1}$tsggukasyan,$^{2}$garnik.davtyan,$^{3}$karavet,$^{4}$ivan.andrianov\}@ispras.ru
}
}

\maketitle

\begin{abstract}
In this work, we tackle the problem of Armenian named entity recognition, providing silver- and gold-standard datasets as well as establishing baseline results on popular models. We present a 163000-token named entity corpus automatically generated and annotated from Wikipedia, and another 53400-token corpus of news sentences with manual annotation of people, organization and location named entities. The corpora were used to train and evaluate several popular named entity recognition models. Alongside the datasets, we release 50-, 100-, 200-, 300-dimensional GloVe word embeddings trained on a collection of Armenian texts from Wikipedia, news, blogs, and encyclopedia.

\end{abstract}


\section{Introduction}

Named entity recognition is an important task of natural language processing, featuring in many popular text processing toolkits. This area of natural language processing has been actively studied in the latest decades and the advent of deep learning reinvigorated the research on more effective and accurate models. However, most of existing approaches require large annotated corpora. To the best of our knowledge, no such work has been done for the Armenian language, and in this work we address several problems, including the creation of a corpus for training machine learning models, the development of gold-standard test corpus and evaluation of the effectiveness of established approaches for the Armenian language.

Considering the cost of creating manually annotated named entity corpus, we focused on alternative approaches. Lack of named entity corpora is a common problem for many languages, thus bringing the attention of many researchers around the globe. Projection based transfer schemes have been shown to be very effective (e.g. \cite{yarowski}, \cite{zitouni}, \cite{ehrmann}), using resource-rich language's corpora to generate annotated data for the low-resource language. In this approach, the annotations of high-resource language are projected over the corresponding tokens of the parallel low-resource language's texts. This strategy can be applied for language pairs that have parallel corpora. However, that approach would not work for Armenian as we did not have access to sufficiently large parallel corpus with a resource-rich language.

Another popular approach is using Wikipedia. Klesti Hoxha and Artur Baxhaku employ gazetteers extracted from Wikipedia to generate an annotated corpus for Albanian\cite{albanian}, and Weber and P\"otzl propose a rule-based system for German that leverages the information from Wikipedia\cite{german}. However, the latter relies on external tools such as part-of-speech taggers, making it nonviable for the Armenian language.

Nothman et al. generated a silver-standard corpus for 9 languages by extracting Wikipedia article texts with outgoing links and turning those links into named entity annotations based on the target article's type\cite{nothman}. Sysoev and Andrianov used a similar approach for the Russian language\cite{sysoev}\cite{texterra}. Based on its success for a wide range of languages, our choice fell on this model to tackle automated data generation and annotation for the Armenian language.

Aside from the lack of training data, we also address the absence of a benchmark dataset of Armenian texts for named entity recognition. We propose a gold-standard corpus with manual annotation of CoNLL named entity categories: person, location, and organization \cite{conll02}\cite{conll03}, hoping it will be used to evaluate future named entity recognition models.

Furthermore, popular entity recognition models were applied to the mentioned data in order to obtain baseline results for future research in the area. Along with the datasets, we developed GloVe\cite{glove} word embeddings to train and evaluate the deep learning models in our experiments.

The contributions of this work are (i) the silver-standard training corpus, (ii) the gold-standard test corpus, (iii) GloVe word embeddings, (iv) baseline results for 3 different models on the proposed benchmark data set. All aforementioned resources are available on GitHub\footnote{https://github.com/ispras-texterra/pioner}.

\section{Automated training corpus generation}

We used Sysoev and Andrianov's modification of the Nothman et al. approach to automatically generate data for training a named entity recognizer. This approach uses links between Wikipedia articles to generate sequences of named-entity annotated tokens.

\subsection{Dataset extraction}\label{approach}

\begin{figure}
\caption{Steps of automatic dataset extraction from Wikipedia}

\centering
    \begin{tikzpicture}[every node/.style={rectangle, fill=blue!20,minimum width=80mm,minimum height=7mm,draw,rounded corners=.6ex}]
\tikzset{arrow/.style={color=black,draw=blue,-latex,font=\fontsize{8}{8}\selectfont}}

 \node (1) at (0,0){Classification of Wikipedia articles into NE types};
 \node (2) at (0,-1.5){Labelling common article aliases to increase coverage}; 
 \node (3) at (0,-3){Extraction of text fragments with outgoing links}; 
 \node (4) at (0,-4.5){Labelling links according to their target article’s type};
 \node (5) at (0,-6){Adjustment of labeled entities' boundaries}; 


  \draw[arrow] (1) -> (2);%
  \draw[arrow] (2) -> (3);%
  \draw[arrow] (3) -> (4);%
  \draw[arrow] (4) -> (5);%

\end{tikzpicture}
\label{fig:1}
\end{figure}
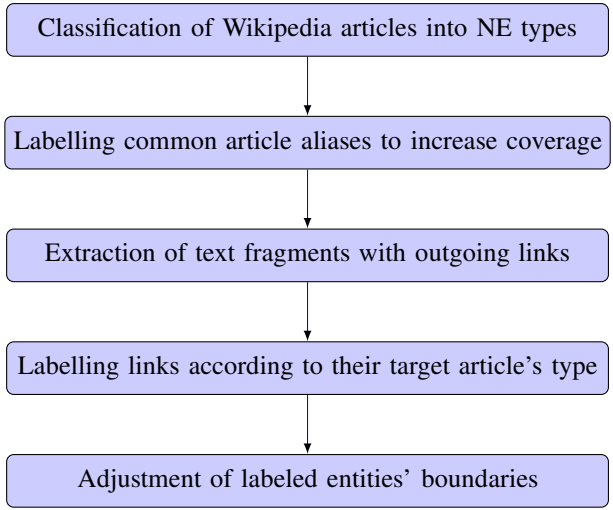

The main steps of the dataset extraction system are described in Figure \ref{fig:1}.

First, each Wikipedia article is assigned a named entity class (e.g. the article \textarmenian{\textsf{Քիմ Քաշքաշյան}} (Kim Kashkashian) is classified as \textsf{PER} (person), \textarmenian{\textsf{Ազգերի լիգա}}(League of Nations) as \textsf{ORG} (organization), \textarmenian{\textsf{Սիրիա}}(Syria) as \textsf{LOC} etc). One of the core differences between our approach and Nothman's system is that we do not rely on manual classification of articles and do not use inter-language links to project article classifications across languages. Instead, our classification algorithm uses only an article's Wikidata entry's first \textsf{instance of} label's parent \textsf{subclass of} labels, which are, incidentally, language independent and thus can be used for any language.

Then, outgoing links in articles are assigned the article's type they are leading to. Sentences are included in the training corpus only if they contain at least one named entity and all contained capitalized words have an outgoing link to an article of known type. Since in Wikipedia articles only the first mention of each entity is linked, this approach becomes very restrictive and in order to include more sentences, additional links are inferred. This is accomplished by compiling a list of common aliases for articles corresponding to named entities, and then finding text fragments matching those aliases to assign a named entity label. An article's aliases include its title, titles of disambiguation pages with the article, and texts of links leading to the article (e.g. \textarmenian{Լենինգրադ} (Leningrad), \textarmenian{Պետրոգրադ} (Petrograd), \textarmenian{Պետերբուրգ} (Peterburg) are aliases for \textarmenian{Սանկտ Պետերբուրգ} (Saint Petersburg)). The list of aliases is compiled for all \textsf{PER}, \textsf{ORG}, \textsf{LOC} articles.

After that, link boundaries are adjusted by removing the labels for expressions in parentheses, the text after a comma, and in some cases breaking into separate named entities if the linked text contains a comma. For example, \textsf{[LOC \textarmenian{Աբովյան (քաղաք)}]} (Abovyan (town)) is reworked into \textsf{[LOC \textarmenian{Աբովյան}] \textarmenian{(քաղաք)}}.  

\begin{table*}[!htb]
\caption{The mapping of Wikidata subclass of values to named entity types}
\begin{center}
\begin{tabular}{|p{10cm}|c|}

\hline
\textbf{subclass of} & \textbf{NE type} \\

\hline
company, business enterprise, company, juridical person, air carrier, political organization, government organization, secret service, political party, international organization, alliance, armed organization, higher education institution, educational institution, university, educational organization, school, fictional magic school, broadcaster, newspaper, periodical literature, religious organization, football club, sports team, musical ensemble, music organisation, vocal-musical ensemble, sports organization, criminal organization, museum of culture, scientific organisation, non-governmental organization, nonprofit organization, national sports team, legal person, scholarly publication, academic journal, association, band, sports club, institution, medical facility
& ORG \\
\hline
state, disputed territory, country, occupied territory, political territorial entity, city, town, village, rural area, rural settlement, urban-type settlement, geographical object, geographic location, geographic region, community, administrative territorial entity, former administrative territorial entity, human settlement, county, province, federated state, district, county-equivalent, municipal formation, raion, nahiyah, mintaqah, muhafazah, realm, principality, historical country, watercourse, lake, sea, still waters, body of water, landmass, minor planet, landform, natural geographic object, mountain range, mountain, protected area, national park, geographic region, geographic location, arena, bridge, airport, stadium, performing arts center, public building, venue, sports venue, church, temple, place of worship, retail building
& LOC \\
\hline
person, fictional character, fictional humanoid,
human who may be fictional, given name, fictional human, magician in fantasy & PER \\
\hline

\end{tabular}
\label{table:mapping}
\end{center}
\end{table*}

\subsection{Using Wikidata to classify Wikipedia}

Instead of manually classifying Wikipedia articles as it was done in Nothman et al., we developed a rule-based classifier that used an article's Wikidata \textsf{instance of} and \textsf{subclass of} attributes to find the corresponding named entity type.

The classification could be done using solely \textsf{instance of} labels, but these labels are unnecessarily specific for the task and building a mapping on it would require a more time-consuming and meticulous work. Therefore, we classified articles based on their first \textsf{instance of} attribute's \textsf{subclass of} values. Table \ref{table:mapping} displays the mapping between these values and named entity types. Using higher-level \textsf{subclass of} values was not an option as their values often were too general, making it impossible to derive the correct named entity category.

\subsection{Generated data}

Using the algorithm described above, we generated 7455 annotated sentences with 163247 tokens based on 20 February 2018 dump of Armenian Wikipedia.

The generated data is still significantly smaller than the manually annotated corpora from CoNLL 2002 and 2003. For comparison, the train set of English CoNLL 2003 corpus contains 203621 tokens and the German one 206931, while the Spanish and Dutch corpora from CoNLL 2002 respectively 273037 and 218737 lines. The smaller size of our generated data can be attributed to the strict selection of candidate sentences as well as simply to the relatively small size of Armenian Wikipedia.

The accuracy of annotation in the generated corpus heavily relies on the quality of links in Wikipedia articles. During generation, we assumed that first mentions of all named entities have an outgoing link to their article, however this was not always the case in actual source data and as a result the train set contained sentences where not all named entities are labeled. Annotation inaccuracies also stemmed from wrongly assigned link boundaries (for example, in Wikipedia article \textarmenian{Արթուր Ուելսլի Վելինգթոն} (Arthur Wellesley) there is a link to the Napoleon article with the text "\textarmenian{է Նապոլեոնը}" ("Napoleon is"), when it should be "\textarmenian{Նապոլեոնը}" ("Napoleon")). Another kind of common annotation errors occurred when a named entity appeared inside a link not targeting a \textsf{LOC}, \textsf{ORG}, or \textsf{PER} article (e.g. "\textarmenian{ԱՄՆ նախագահական ընտրություններում}" ("USA presidential elections") is linked to the article \textarmenian{ԱՄՆ նախագահական ընտրություններ 2016} (United States presidential election, 2016) and as a result [\textsf{LOC \textarmenian{ԱՄՆ}}] (USA) is lost).

\section{Test dataset}\label{experiments}

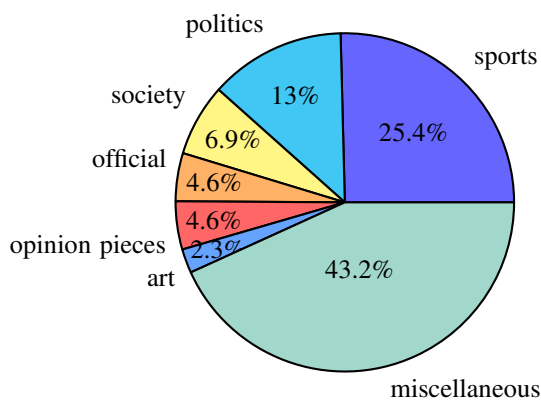
\begin{figure}[h]
\caption{Topics distribution in the gold-standard corpus}
\begin{tikzpicture}
\begin{scope}[scale=.75]
    \pie{25.4/sports, 13/politics, 6.9/society, 4.6/official, 4.6/opinion pieces, 2.3/art, 43.2/miscellaneous }
\end{scope}
\end{tikzpicture}
\label{fig:topicpie}
\end{figure}

\begin{figure}[h]
\caption{Distribution of examples by location in the gold-standard corpus}
\centering
\begin{tikzpicture}
\begin{scope}[scale=.75]
    \pie{40/local, 13.5/world, 9.6/regional, 36.9/miscellaneous }
\end{scope}
\end{tikzpicture}
\label{fig:locpie}
\end{figure}
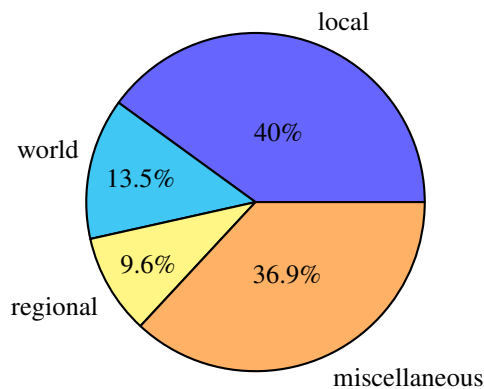

In order to evaluate the models trained on generated data, we manually annotated a named entities dataset comprising 53453 tokens and 2566 sentences selected from over 250 news texts from ilur.am\footnote{http://ilur.am/news/newsline.html}. This dataset is comparable in size with the test sets of other languages (Table \ref{table:testsets}). Included sentences are from political, sports, local and world news (Figures \ref{fig:topicpie}, \ref{fig:locpie}), covering the period between August 2012 and July 2018. The dataset provides annotations for 3 popular named entity classes: people (\textsf{PER}), organizations (\textsf{ORG}), and locations (\textsf{LOC}), and is released in CoNLL03 format with IOB tagging scheme. Tokens and sentences were segmented according to the UD standards for the Armenian language\cite{armtdp}.

\begin{table}[H]
\caption{Comparison of Armenian, English, German, Spanish and Russian test sets: sentence, token, and named entity counts}
\begin{center}
\begin{tabular}{|l|c|c|c|c|}

\hline
\textbf{Test set} & \textbf{Tokens} & \textbf{LOC} & \textbf{ORG} & \textbf{PER}\\

\hline
Armenian & 53453 & 1306 & 1337 & 1274 \\
\hline
English CoNLL03 & 46435 & 1668 & 1661 & 1617 \\
\hline
German CoNLL03 & 51943 & 1035 & 773 & 1195 \\
\hline
Spanish CoNLL02 & 51533 & 1084 & 1400 & 735 \\
\hline
Russian factRuEval-2016 & 59382 & 1239 & 1595 & 1353 \\
\hline

\end{tabular}
\label{table:testsets}
\end{center}
\end{table}

During annotation, we generally relied on categories and guidelines assembled by BBN Technologies for TREC 2002 question answering track\footnote{https://catalog.ldc.upenn.edu/docs/LDC2005T33/BBN-Types-Subtypes.html}. Only named entities corresponding to BBN's \textsf{person name} category were tagged as \textsf{PER}. Those include proper names of people, including fictional people, first and last names, family names, unique nicknames. Similarly, \textsf{organization name} categories, including company names, government agencies, educational and academic institutions, sports clubs, musical ensembles and other groups, hospitals, museums, newspaper names, were marked as \textsf{ORG}. However, unlike BBN, we did not mark adjectival forms of organization names as named entities. BBN's \textsf{gpe name}, \textsf{facility name}, \textsf{location name} categories were combined and annotated as \textsf{LOC}.

We ignored entities of other categories (e.g. works of art, law, or events), including those cases where an \textsf{ORG}, \textsf{LOC} or \textsf{PER} entity was inside an entity of extraneous type (e.g. \textarmenian{ՀՀ} (RA) in \textarmenian{ՀՀ Քրեական Օրենսգիրք} (RA Criminal Code) was not annotated as \textsf{LOC}).

Quotation marks around a named entity were not annotated unless those quotations were a part of that entity's full official name (e.g. \textarmenian{«Նաիրիտ գործարան» ՓԲԸ} ("Nairit Plant" CJSC)).

Depending on context, metonyms such as \textarmenian{Կրեմլ} (Kremlin), \textarmenian{Բաղրամյան 26} (Baghramyan 26) were annotated as \textsf{ORG} when referring to respective government agencies. Likewise, country or city names were also tagged as \textsf{ORG} when referring to sports teams representing them.

\section{Word embeddings}

Apart from the datasets, we also developed word embeddings for the Armenian language, which we used in our experiments to train and evaluate named entity recognition algorithms. Considering their ability to capture semantic regularities, we used GloVe to train word embeddings. We assembled a dataset of Armenian texts containing 79 million tokens from the articles of Armenian Wikipedia, The Armenian Soviet Encyclopedia, a subcorpus of Eastern Armenian National Corpus\cite{eanc}, over a dozen Armenian news websites and blogs. Included texts covered topics such as economics, politics, weather forecast, IT, law, society and politics, coming from non-fiction as well as fiction genres.

Similar to the original embeddings published for the English language, we release 50-, 100-, 200- and 300-dimensional word vectors for Armenian with a vocabulary size of 400000. Before training, all the words in the dataset were lowercased. For the final models we used the following training hyperparameters: 15 window size and 20 training epochs.


\section{Experiments}

In this section we describe a number of experiments targeted to compare the performance of popular named entity recognition algorithms on our data. We trained and evaluated Stanford NER\footnote{https://nlp.stanford.edu/software/CRF-NER.shtml}, spaCy 2.0\footnote{https://spacy.io/}, and a recurrent model similar to \cite{lample},\cite{mahovy} that uses bidirectional LSTM cells for character-based feature extraction and CRF, described in Guillaume Genthial's \textit{Sequence Tagging with Tensorflow} blog post \cite{genthial}.

\subsection{Models}

Stanford NER is conditional random fields (CRF) classifier based on lexical and contextual features such as the current word, character-level n-grams of up to length 6 at its beginning and the end, previous and next words, word shape and sequence features \cite{stanfordner}.

spaCy 2.0 uses a CNN-based transition system for named entity recognition. For each token, a Bloom embedding is calculated based on its lowercase form, prefix, suffix and shape, then using residual CNNs, a contextual representation of that token is extracted that potentially draws information from up to 4 tokens from each side \cite{strubell}. Each update of the transition system's configuration is a classification task that uses the contextual representation of the top token on the stack, preceding and succeeding tokens, first two tokens of the buffer, and their leftmost, second leftmost, rightmost, second rightmost children. The valid transition with the highest score is applied to the system. This approach reportedly performs within 1\% of the current state-of-the-art for English \footnote{https://spacy.io/usage/v2\#features-models}. In our experiments, we tried out 50-, 100-, 200- and 300-dimensional pre-trained GloVe embeddings. Due to time constraints, we did not tune the rest of hyperparameters and used their default values.

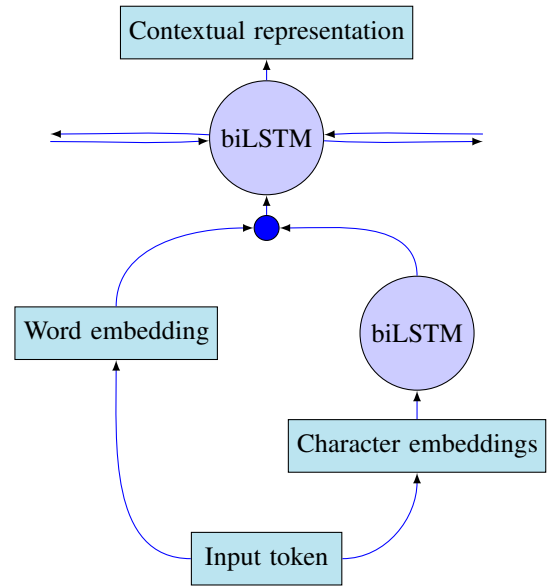
\begin{figure}
\caption{The neural architecture for extracting contextual representations of tokens}

\centering
    \begin{tikzpicture}[
    process/.style={circle, fill=blue!20,minimum width=10mm,minimum height=7mm,draw},
    data/.style={rectangle, fill=SkyBlue!40,minimum width=20mm,minimum height=7mm,draw},
    dot/.style={circle, fill=blue!100,minimum width=1mm,minimum height=1mm,draw}]

\tikzset{arrow/.style={color=black,draw=blue,-latex,font=\fontsize{8}{8}\selectfont}}

 \node[data] (input) at (1,0){ Input token};
 \node[data] (word) at (-1,3) {Word embedding}; 
 \node[data] (char) at (3,1.5) {Character embeddings}; 
 \node[process] (charbilstm) at (3,3) {biLSTM}; 
 \node[dot] (join) at (1,4.4) {}; 
 \node[process] (bilstm) at (1,5.6){biLSTM};
 \node[blue] (lefttop) at (-2,5.65) {};
 \node[blue] (leftbtm) at (-2,5.55) {};
 \node[blue] (righttop) at (4,5.65) {}; 
 \node[blue] (rightbtm) at (4,5.55) {};
 \node[data] (crf) at (1,7){Contextual representation};


  \draw[arrow] (input) to [out=180,in=-90] (word);%
  \draw[arrow] (input) to [out=0,in=-90] (char);%
  \draw[arrow] (char) to [out=90,in=-90] (charbilstm);%
  \draw[arrow] (word) to [out=90,in=180] (join);%
  \draw[arrow] (charbilstm) to [out=90,in=0] (join);%
  \draw[arrow] (join) to [out=90,in=-90] (bilstm);%
  \draw[arrow] (leftbtm) to [out=0,in=183] (bilstm);%
  \draw[arrow] (bilstm) to [out=-3,in=180] (rightbtm);%
  \draw[arrow] (bilstm) to [out=177,in=0] (lefttop);%
  \draw[arrow] (righttop) to [out=180,in=3] (bilstm);%
  \draw[arrow] (bilstm) -> (crf);%

\end{tikzpicture}
\label{fig:bilstm}
\end{figure}

The main model that we focused on was the recurrent model with a CRF top layer, and the above-mentioned methods served mostly as baselines. The distinctive feature of this approach is the way contextual word embeddings are formed. For each token separately, to capture its word shape features, character-based representation is extracted using a bidirectional LSTM \cite{ling}. This representation gets concatenated with a distributional word vector such as GloVe, forming an intermediate word embedding. Using another bidirectional LSTM cell on these intermediate word embeddings, the contextual representation of tokens is obtained (Figure \ref{fig:bilstm}). Finally, a CRF layer labels the sequence of these contextual representations. In our experiments, we used Guillaume Genthial's implementation\footnote{https://github.com/guillaumegenthial/sequence\_tagging} of the algorithm. We set the size of character-based biLSTM to 100 and the size of second biLSTM network to 300.

\subsection{Evaluation}

Experiments were carried out using IOB tagging scheme, with a total of 7 class tags: O, B-PER, I-PER, B-LOC, I-LOC, B-ORG, I-ORG. 

We randomly selected 80\% of generated annotated sentences for training and used the other 20\% as a development set. The models with the best F1 score on the development set were tested on the manually annotated gold dataset.

\begin{table*}[htb]
\caption{Evaluation results for named entity recognition algorithms}
\begin{center}
\begin{tabular}{|l|c|c|c|c|c|c|}

\hline
\multicolumn{1}{|c|}{\multirow{2}*{\textbf{Algorithm}}}
& \multicolumn{3}{|c|}{\textbf{dev}}
& \multicolumn{3}{|c|}{\textbf{test}} \\
\cline{2-7}

& \textbf{Precision} & \textbf{Recall} & \textbf{F1} & \textbf{Precision} & \textbf{Recall} & \textbf{F1}\\

\hline
Stanford NER & 76.86 & 70.62 & 73.61 & \textbf{78.46} & 46.52 & 58.41 \\
\hline
spaCy 2.0 & 68.19 & 71.86 & 69.98 & 64.83 & \textbf{55.77} & 59.96 \\
\hline
Char-biLSTM+biLSTM+CRF & \textbf{77.21} & \textbf{74.81} & \textbf{75.99} & 73.27 & 54.14 & \textbf{62.23} \\
\hline

\end{tabular}
\label{table:results}
\end{center}
\end{table*}

\begin{table*}[htb]
\caption{Evaluation results for Char-biLSTM+biLSTM+CRF}
\begin{center}
\begin{tabular}{|l|c|c|c|c|c|c|c|c|c|c|c|c|}

\hline
\multicolumn{1}{|c|}{\multirow{2}*{\textbf{Word embeddings}}}
& \multicolumn{2}{|c|}{\textbf{train embeddings=False}}
& \multicolumn{2}{|c|}{\textbf{train embeddings=True}} \\
\cline{2-5}

& \textbf{dev F1} & \textbf{test F1} & \textbf{dev F1} & \textbf{test F1}\\
\cline{2-5}

\hline
GloVe (dim=50) & 74.18 & 56.46 & \textbf{75.99} & \textbf{62.23}\\
\hline
GloVe (dim=100) & 73.94 & 58.52 & 74.83 & 61.54\\
\hline
GloVe (dim=200) & \textbf{75.00} & 58.37 & 74.97 & 59.78\\
\hline
GloVe (dim=300) & 74.21 & \textbf{59.66} & 74.75 & 59.92\\
\hline

\end{tabular}
\label{table:genthial}
\end{center}
\end{table*}

\begin{table*}[htb]
\caption{Evaluation results for spaCy 2.0 NER}
\begin{center}
\begin{tabular}{|l|c|c|c|c|c|c|}

\hline
\multicolumn{1}{|c|}{\multirow{2}*{\textbf{Word embeddings}}}
& \multicolumn{3}{|c|}{\textbf{dev}}
& \multicolumn{3}{|c|}{\textbf{test}} \\
\cline{2-7}

& \textbf{Precision} & \textbf{Recall} & \textbf{F1} & \textbf{Precision} & \textbf{Recall} & \textbf{F1} \\

\hline
GloVe (dim=50) & 69.31 & 71.26 & 70.27 & 66.52 & 51.72 & 58.20 \\
\hline
GloVe (dim=100) & \textbf{70.12} & \textbf{72.91} & \textbf{71.49} & 66.34 & 53.35 & 59.14\\
\hline
GloVe (dim=200) & 68.19 & 71.86 & 69.98 & 64.83 & \textbf{55.77} & \textbf{59.96} \\
\hline
GloVe (dim=300) & 70.08 & 71.80 & 70.93 & \textbf{66.61} & 52.94 & 59.00 \\
\hline

\end{tabular}
\label{table:spacy}
\end{center}
\end{table*}

\section{Discussion}\label{discussion}

Table \ref{table:results} shows the average scores of evaluated models. The highest F1 score was achieved by the recurrent model using a batch size of 8 and Adam optimizer with an initial learning rate of 0.001. Updating word embeddings during training also noticeably improved the performance. GloVe word vector models of four different sizes (50, 100, 200, and 300) were tested, with vectors of size 50 producing the best results (Table \ref{table:genthial}).

For spaCy 2.0 named entity recognizer, the same word embedding models were tested. However, in this case the performance of 200-dimensional embeddings was highest (Table \ref{table:spacy}). Unsurprisingly, both deep learning models outperformed the feature-based Stanford recognizer in recall, the latter however demonstrated noticeably higher precision. 

It is clear that the development set of automatically generated examples was not an ideal indicator of models' performance on gold-standard test set. Higher development set scores often led to lower test scores as seen in the evaluation results for spaCy 2.0 and Char-biLSTM+biLSTM+CRF (Tables \ref{table:spacy} and \ref{table:genthial}). Analysis of errors on the development set revealed that many were caused by the incompleteness of annotations, when named entity recognizers correctly predicted entities that were absent from annotations (e.g. [\textarmenian{Խ{Ս}{Հ}Մ-ի} \textsf{LOC}] (USSR's), [\textarmenian{Դինամոն} \textsf{ORG}] (the\_Dinamo), [\textarmenian{Պիրենեյան թերակղզու} \textsf{LOC}] (Iberian Peninsula's) etc). Similarly, the recognizers often correctly ignored non-entities that are incorrectly labeled in data (e.g. [\textarmenian{օսմանների} \textsf{PER}], [\textarmenian{կոնսերվատորիան} \textsf{ORG}] etc). 

Generally, tested models demonstrated relatively high precision of recognizing tokens that started named entities, but failed to do so with descriptor words for organizations and, to a certain degree, locations. The confusion matrix for one of the trained recurrent models illustrates that difference (Table \ref{table:confusion}). This can be partly attributed to the quality of generated data: descriptor words are sometimes superfluously labeled (e.g. [\textarmenian{Հավայան կղզիների տեղաբնիկները} \textsf{LOC}] (the indigenous people of Hawaii)), which is likely caused by the inconsistent style of linking in Armenian Wikipedia (in the article \textarmenian{ԱՄՆ մշակույթ} (Culture of the United States), its linked text fragment "\textarmenian{Հավայան կղզիների տեղաբնիկները}" ("the indigenous people of Hawaii") leads to the article \textarmenian{Հավայան կղզիներ}(Hawaii)).

\begin{table}[htb]

\caption{Confusion matrix on the development set}
\resizebox{\columnwidth}{!}{
\begin{tabular}{cc|ccccccc}
\multicolumn{2}{c}{}
            &   \multicolumn{7}{c}{Predicted} \\
    &    &  O &  B-PER  &  B-ORG  &  B-LOC  &  I-ORG  &  I-PER  &   I-LOC  \\ 
    \cline{2-9}
\multirow{7}{*}{\rotatebox[origin=c]{90}{Actual}}
    & O  & 26707&   100&    57&   249&   150&    78&   129\\
    & B-PER  &   107&   712&     6&    32&     2&     4&     0\\
    & B-ORG &    93&     6&   259&    58&     8&     0&     0\\
    & B-LOC  &   226&    25&    32&  1535&     5&     3&    20\\
    & I-ORG  &    67&     1&     5&     3&   289&     3&    19\\
    & I-PER  &    46&     5&     0&     1&     6&   660&     8\\
    & I-LOC  &   145&     0&     1&    13&    45&    11&   597\\
    \cline{2-9}
    & Precision (\%)  &   97.5&     83.86&     71.94&    81.17&    57.23&    86.95&   77.23\\
\end{tabular}
}
\label{table:confusion}
\end{table}

\section{Conclusion}\label{conclusion}

We release two named-entity annotated datasets for the Armenian language: a silver-standard corpus for training NER models, and a gold-standard corpus for testing. It is worth to underline the importance of the latter corpus, as we aim it to serve as a benchmark for future named entity recognition systems designed for the Armenian language. Along with the corpora, we publish GloVe word vector models trained on a collection of Armenian texts.

Additionally, to establish the applicability of Wikipedia-based approaches for the Armenian language, we provide evaluation results for 3 different named entity recognition systems trained and tested on our datasets. The results reinforce the ability of deep learning approaches in achieving relatively high recall values for this specific task, as well as the power of using character-extracted embeddings alongside conventional word embeddings.

There are several avenues of future work. Since Nothman et al. 2013, more efficient methods of exploiting Wikipedia have been proposed, namely WiNER \cite{winer}, which could help increase both the quantity and quality of the training corpus. Another potential area of work is the further enrichment of the benchmark test set with additional annotation of other classes such as \textsf{MISC} or more fine-grained types (e.g. \textsf{CITY}, \textsf{COUNTRY}, \textsf{REGION} etc instead of \textsf{LOC}).

\end{document}